 \documentclass[smallabstract,smallcaptions]{dccpaper}

\usepackage{epsfig}
\usepackage{amsmath}
\usepackage{amssymb}
\usepackage{color}
\usepackage{url}
\usepackage{multirow}
\usepackage{multicol}
\usepackage{algorithm}
\usepackage{algorithmic}
\usepackage{comment}

\newlength{\figurewidth}
\newlength{\smallfigurewidth}
\def\hms{\hspace{-0.25em}}

\begin{document}

\title
{\large
\textbf{A DenseNet Based Approach for Multi-Frame In-Loop Filter in HEVC}
}

\author{%
Tianyi Li$^{\ast}$, Mai Xu$^{\ast}$, Ren Yang$^{\ast}$ and Xiaoming Tao$^{\dag}$\\[0.5em]
{\small\begin{minipage}{\linewidth}\begin{center}
			\begin{tabular}{c}
				$^{\ast}$School of Electronic and Information Engineering,\\ Beihang University, Beijing, 100191, China\\ 
				$^{\dag}$ Department of Electronic Engineering,\\ Tsinghua University, Beijing, 100084, China\\
				\url{maixu@buaa.edu.cn} (Corresponding author: Mai Xu) 
			\end{tabular}
		\end{center}\end{minipage}}
	}

\maketitle
\thispagestyle{empty}

\begin{abstract}
High efficiency video coding (HEVC) has brought outperforming efficiency for video compression. 
To reduce the compression artifacts of HEVC, we propose a DenseNet based approach as the in-loop filter of HEVC, which leverages multiple adjacent frames to enhance the quality of each encoded frame. 
Specifically, the higher-quality frames are found by a reference frame selector (RFS).
Then, a deep neural network for multi-frame in-loop filter (named MIF-Net) is developed to enhance the quality of each encoded frame by utilizing the spatial information of this frame and the temporal information of its neighboring higher-quality frames. 
The MIF-Net is built on the recently developed DenseNet, benefiting from the improved generalization capacity and computational efficiency.
Finally, experimental results verify the effectiveness of our multi-frame in-loop filter, outperforming the HM baseline and other state-of-the-art approaches.

\end{abstract}

\vspace{-0.25em}
\section{Introduction}
\vspace{-0.25em}

The high efficiency video coding (HEVC) standard \cite{Sullivan12TCSVT} developed by the Joint Collaborate Team on Video Coding (JCT-VC) has brought outperforming efficiency for video compression. 
However, various artifacts (e.g., blocking, blurring and ringing artifacts) also exist in compressed videos, mainly resulting from the block-wise prediction and quantization with limited precision. To alleviate such artifacts, in-loop filters were adopted for enhancing the quality of each encoded frame and providing higher-quality reference for its successive frames. Consequently, the coding efficiency can be further improved by adopting the in-loop filters.

In total, three built-in in-loop filters were proposed for standard HEVC, including deblocking filter (DBF) \cite{Norkin12TCSVT}, sample adaptive offset (SAO) filter \cite{Fu12TCSVT} and adaptive loop filter (ALF) \cite{Tsai13JSTSP}. 
Specifically, DBF is firstly used to remove the blocking artifacts. Then, the SAO filter reduces distortion by adding an adaptive offset to each sample. Afterwards, ALF minimizes the distortion based on Wiener filter. However, ALF introduces heavy bit-rate overhead and it has not been adopted in the final version of HEVC. 
Besides the built-in in-loop filters for HEVC, various heuristic and learning-based methods have also been proposed. In heuristic methods, some prior knowledge of video coding is utilized to build a statistical model of compression artifacts, and then each frame is enhanced based on the model. 
For example, Matsumura \textit{et al}. \cite{Matsumura13} utilized the weighted mean of non-local similar frame patches for artifact reduction. 
Zhang \textit{et al}. \cite{Zhang17TCSVT} attached a low-rank constraint on each matrix formed by a patch group, and then established an adaptive soft-thresholding model to achieve sparse representation.
More recently, deep learning has been successfully employed in many areas about data compression, such as video coding \cite{Xu18TIP} and quality enhancement \cite{Yang18TCSVT}. Also, learning-based methods have further improved the performance of in-loop filtering. 
Among them, Meng \textit{et al}. \cite{MengDCC2018} developed a multi-channel long-short-term dependency residual network (MLSDRN) for mapping a distorted frame to the raw frame, inserted between DBF and SAO.
Zhang \textit{et al}. \cite{Zhang18TIP} proposed a residual highway CNN (RHCNN) based on the ResNet \cite{He2016CVPR_ResNet}, implemented after the standard SAO. 
However, none of the above learning-based methods has employed multiple frames for in-loop filtering in HEVC. Typically, the high fluctuation of visual quality exists across the encoded frames, and thus a low-quality frame can be enhanced by referring to its adjacent higher-quality frames. 

Based on deep learning, this paper develops a multi-frame in-loop filter (MIF) for HEVC, replacing the original DBF and SAO. 
Specifically, we first exploit the quality fluctuation of encoded frames via designing a reference frame selector (RFS) to find reference frames for an unfiltered reconstructed frame (URF), based on frame quality and content similarity. 
If RFS provides sufficient reference frames, the URF flows through a deep neural network for MIF (named MIF-Net) to utilize both spatial information within one frame and temporal information across multiples frames. 
In the case that no sufficient reference frames are selected by RFS, a simpler deep neural network for in-loop filter (named IF-Net) is used to enhance the URF instead.
Considering the blocking artifacts influenced by the coding tree unit (CTU) partition, the proposed networks are also adaptive to the partition structure, via varying convolutional kernels at different locations of the coding unit (CU) and transform unit (TU) maps. 
Finally, the experimental results show that our approach outperforms other state-of-the-art approaches, with $5.33\%$ and $2.40\%$ saving of the Bj\o{}ntegaard delta bit-rate (BD-BR) over the non-local adaptive loop filter \cite{Zhang17TCSVT} and the RHCNN \cite{Zhang18TIP}, respectively.


\vspace{-0.75em}
\section{Proposed MIF Approach}
\label{sec:method}
\vspace{-0.75em}

\vspace{-0.5em}
\subsection{Framework}
\label{sec:framework}
\vspace{-0.5em}

\begin{figure*}
	\centering
	\includegraphics[width=0.9\linewidth]{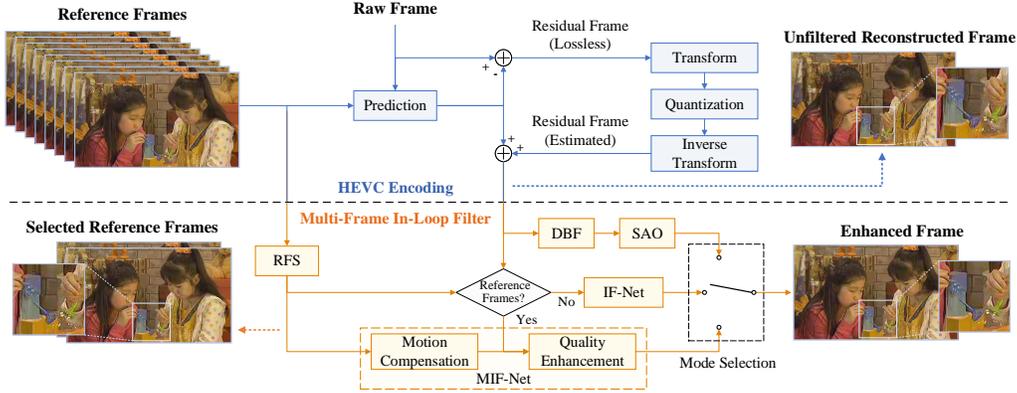}
	\vspace{-1.75em}
	\caption{Framework of the proposed MIF.}
	\label{fig:framework}
	\vspace{-2em}
\end{figure*}

The framework of our MIF approach is illustrated in Figure \ref{fig:framework}. 
In the standard HEVC, each raw frame is encoded through intra/inter-mode prediction, discrete transform and quantization. Then, the predicted frame and the residual frame form a URF. Subsequently, the URF is filtered with DBF and SAO for quality enhancement.
Different from the standard HEVC, we propose a deep-learning-based in-loop filter to enhance the URF, leveraging information from its neighboring frames.
First, RFS selects high quality and high correlated frames as reference, to be introduced in Section \ref{sec:rfs}.
Next, one of the two possible filtering modes is adopted to the URF, as described below.
\begin{itemize}
	\item
	\vspace{-0.5em}
	\textbf{Mode 1: MIF-Net.}
	Assume that $M$ reference frames are needed in MIF-Net. If RFS selects at least $M$ frames, the URF is processed by MIF-Net to generate an enhanced frame. 
	In MIF-Net, each reference frame is first aligned with the URF in terms of content, with a motion compensation network. 
	Then, both aligned reference frames and the URF are fed into a quality enhancement network to output the reconstructed frame. 
	\item
	\vspace{-0.5em}
	\textbf{Mode 2: IF-Net.}
	If no enough reference frames are found for the URF, IF-Net is adopted instead for quality enhancement.
	In contrast to MIF-Net, IF-Net takes only the URF as input without any consideration of multiple frames. 
	\vspace{-0.5em}
\end{itemize}
More details about Modes 1 and 2 are presented in Section \ref{sec:net}. 
If MIF-Net or IF-Net fails to improve frame quality, the standard DBF and SAO can also be used as a supplementary mode. 
Finally, the best mode among the three possible choices (i.e., MIF-Net, IF-Net and the standard in-loop filters) is selected as the actual choice, ensuring the overall performance of our approach.

\vspace{-0.5em}
\subsection{Design of RFS}
\label{sec:rfs}
\vspace{-0.5em}

In our approach, RFS selects reference frames for each URF. For the $n$-th URF (denoted as $\mathbf{F}^\mathrm{U}_n$) in a sequence, RFS examines its previous $N$ encoded frames as the reference frame pool, each of which is denoted by $\mathbf{F}^\mathrm{P}_i$ ($n-N\le{}i\le{}n-1$). Afterwards, six metrics reflecting quality difference and content similarity are calculated. 
\begin{itemize}
	\vspace{-0.5em}
	\item
	$\Delta{}\mathrm{PSNR}_{i,n}^\mathrm{Y}$, $\Delta{}\mathrm{PSNR}_{i,n}^\mathrm{U}$ and $\Delta{}\mathrm{PSNR}_{i,n}^\mathrm{V}$: PSNR increment of $\mathbf{F}^\mathrm{P}_i$ over $\mathbf{F}^\mathrm{U}_n$, for the Y, U and V channels, respectively.
	\vspace{-0.5em}
	\item
	$\mathrm{CC}_{i,n}^\mathrm{Y}$, $\mathrm{CC}_{i,n}^\mathrm{U}$ and $\mathrm{CC}_{i,n}^\mathrm{V}$: the correlation coefficient (CC) values of frame content between $\mathbf{F}^\mathrm{P}_i$ and $\mathbf{F}^\mathrm{U}_n$ for the Y, U and V channels, respectively.
	\vspace{-0.5em}
\end{itemize}
Based on the above metrics, the reference frame pool is first divided into valid and invalid reference frames, and then all valid frames are fed into RFS-Net to select $M$ frames in final. 
Specifically, a binary value $V_{i,n}$ represents whether a reference frame from the pool is valid. 
For at least one channel of $\mathbf{F}^\mathrm{P}_i$, if the PSNR increment is positive and the CC value is above a threshold $\tau$, i.e., $V_{i,n}=1$ in (\ref{eq:valid-frame}), $\mathbf{F}^\mathrm{P}_i$ is seen as a valid reference frame.
\vspace{-0.5em}
\begin{equation}\label{eq:valid-frame}
	\! V_{i,n}=\left\{
	\begin{array}{ll}
		\!\! 1,           &\mathrm{if} \!\!\!\! \displaystyle\bigvee_{c\in\{\mathrm{Y,U,V}\}} \!\!\!\!\!\! (\Delta{}\mathrm{PSNR}_{i,n}^{c}>0 \wedge \mathrm{CC}_{i,n}^{c}>\tau) \\  
		\!\! 0,           &       \mathrm{otherwise.}\\
	\end{array} \right.
\end{equation}

If there exist at least $M$ valid reference frames, the six metrics for each valid reference frame form a 6-dimensional vector, and then they are input to a two-layer fully connected network (named RFS-Net\footnote{The 6-dimensional vector flows through two layers, with 12 hidden nodes and 1 output node, respectively. Both layers are activated with parametric rectified linear units (PReLU) \cite{He15ICCV_PReLU}. Note that the samples in one training batch are extracted from the valid reference frames for only one URF, and the output of samples in the same batch are Z-score normalized.}) to output a scalar $\hat{R}_{i,n}$.
The output $\hat{R}_{i,n}$ is a continuous variable representing the potential of $\mathbf{F}^\mathrm{P}_i$ being the reference for $\mathbf{F}^\mathrm{U}_n$.
A larger $\hat{R}_{i,n}$ indicates that $\mathbf{F}^\mathrm{P}_i$ has more potential than other reference frames for enhancing $\mathbf{F}^\mathrm{U}_n$. Here, $\hat{R}_{i,n}$ is the predicted value by RFS-Net, with the corresponding ground-truth denoted by $R_{i,n}$. 
In RFS-Net, the ground-truth $R_{i,n}$ should reflect the quality of a valid reference frame after it is aligned with $\mathbf{F}^\mathrm{U}_n$ via motion compensation. 
To this end, we assign $R_{i,n}$ as the PSNR between the compensated valid reference frame and the $n$-th raw frame (denoted as $\mathbf{F}_n$). In accord with $\hat{R}_{i,n}$, the $R_{i,n}$ is also Z-scored normalized within one training batch. 
After normalization, the $\ell_2$-loss on the whole training batch can be used to measure the difference between $R_{i,n}$ and $\hat{R}_{i,n}$, formulated as
\vspace{-0.25em}
\begin{equation}\label{eq:loss-rfs}
L_\mathrm{RFS}= \!\!\!\! \sum_{\substack{n-N\le{}i\le{}n-1,\:V_{i,n}=1}} \!\!\!\! (R_{i,n}-\hat{R}_{i,n})^2,
\vspace{-0.25em}
\end{equation}
which is optimized by the Adam algorithm \cite{Kingma2014CS_Adam}. 
Using the trained RFS-Net model, the reference potential for all the valid frames can be obtained. Then, RFS selects $M$ frames denoted by $\{\mathbf{F}^\mathrm{R}_{m,n}\}_{m=1}^M$, where the index $m$ indicates that $\mathbf{F}^\mathrm{R}_{m,n}$ is the frame with the $m$-th highest $\hat{R}_{i,n}$ among all valid reference frames. In the exceptional case that the number of valid reference frames is less than $M$, RFS does not work and IF-Net is used to enhance $\mathbf{F}^\mathrm{U}_n$ instead.

\vspace{-0.5em}
\subsection{MIF-Net and IF-Net}
\label{sec:net}
\vspace{-0.5em}

\begin{figure*}
	\centering
	\includegraphics[width=0.85\linewidth]{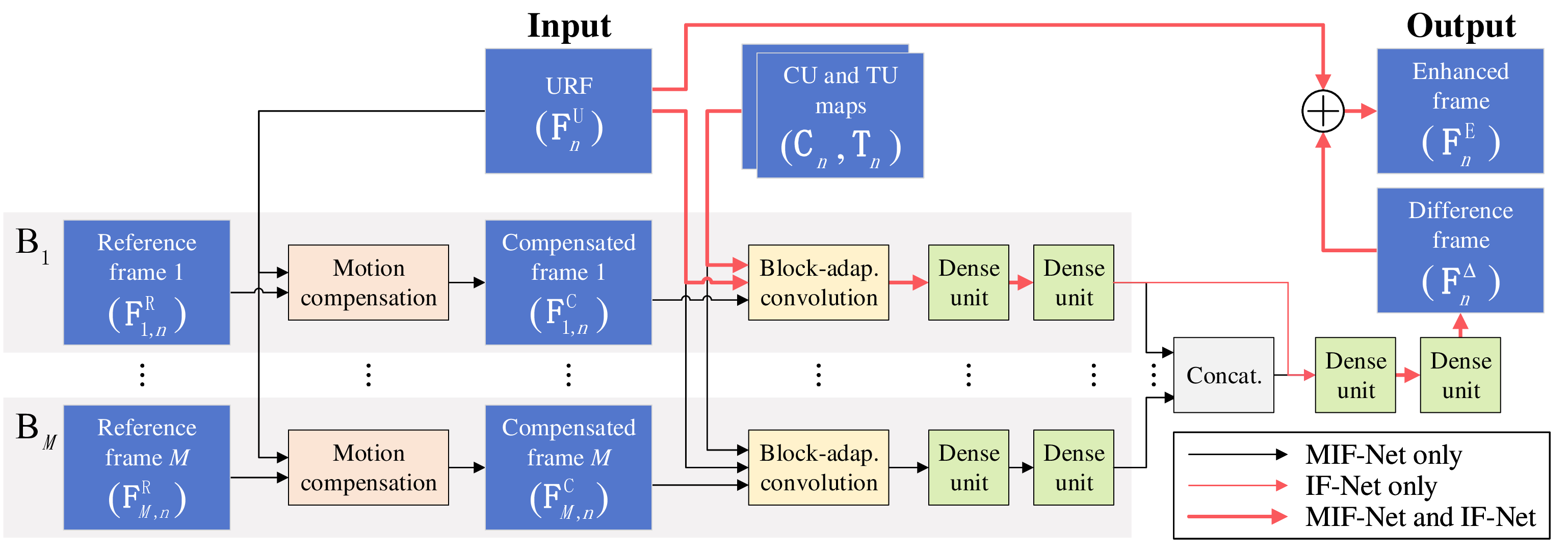}
	\vspace{-2em}
	\caption{Architecture of MIF-Net or IF-Net.}
	\vspace{-2.25em}
	\label{fig:net}
\end{figure*}

This section mainly focuses on the architecture of MIF-Net and its training strategy, and then specifies the difference between IF-Net and MIF-Net.
Figure \ref{fig:net} illustrates the overall architecture of MIF-Net or IF-Net. As shown in this figure, MIF-Net takes a URF $\mathbf{F}^\mathrm{U}_n$ and its $M$ reference frames $\{\mathbf{F}^\mathrm{R}_{m,n}\}_{m=1}^M$ as the input, to generate the enhanced frame $\mathbf{F}^\mathrm{E}_n$ as the output. The information from $M$ parallel branches $\{\mathbf{B}_m\}_{m=1}^M$ is synthesized, with each branch $\mathbf{B}_m$ dealing with the corresponding reference frame $\mathbf{F}^\mathrm{R}_{m,n}$.
In branch $\mathbf{B}_m$, $\mathbf{F}^\mathrm{R}_{m,n}$ is first aligned with $\mathbf{F}^\mathrm{U}_n$ to produce a motion-compensated frame, denoted as $\mathbf{F}^\mathrm{C}_{m,n}$. 
Next, $\mathbf{F}^\mathrm{U}_n$ with $\mathbf{F}^\mathrm{C}_{m,n}$ flows through a novel convolutional layer guided by the CTU partition structure of $\mathbf{F}^\mathrm{U}_n$ (named block-adaptive convolutional layer), to explore low-level features from different sources and merge the features with consideration of the CU and TU partition. 
Then, the low-level features flow through two successive dense units \cite{Huang2017CVPR_DenstNet} to extract more comprehensive features within $\mathbf{B}_m$. Finally, the extracted features from $M$ branches are concatenated together and further processed with two dense units to extract high-level features. For ease of training, the output of the last dense unit (denoted as $\mathbf{F}^\mathrm{\Delta}_n$) is regarded as a difference frame, and the enhanced frame $\mathbf{F}^\mathrm{E}_n$ is the summation of $\mathbf{F}^\mathrm{\Delta}_n$ and $\mathbf{F}^\mathrm{U}_n$.
The details of MIF-Net components are presented in the following.

\begin{figure}
	\begin{minipage}{0.6\linewidth}
		\centering
		\includegraphics[width=90mm]{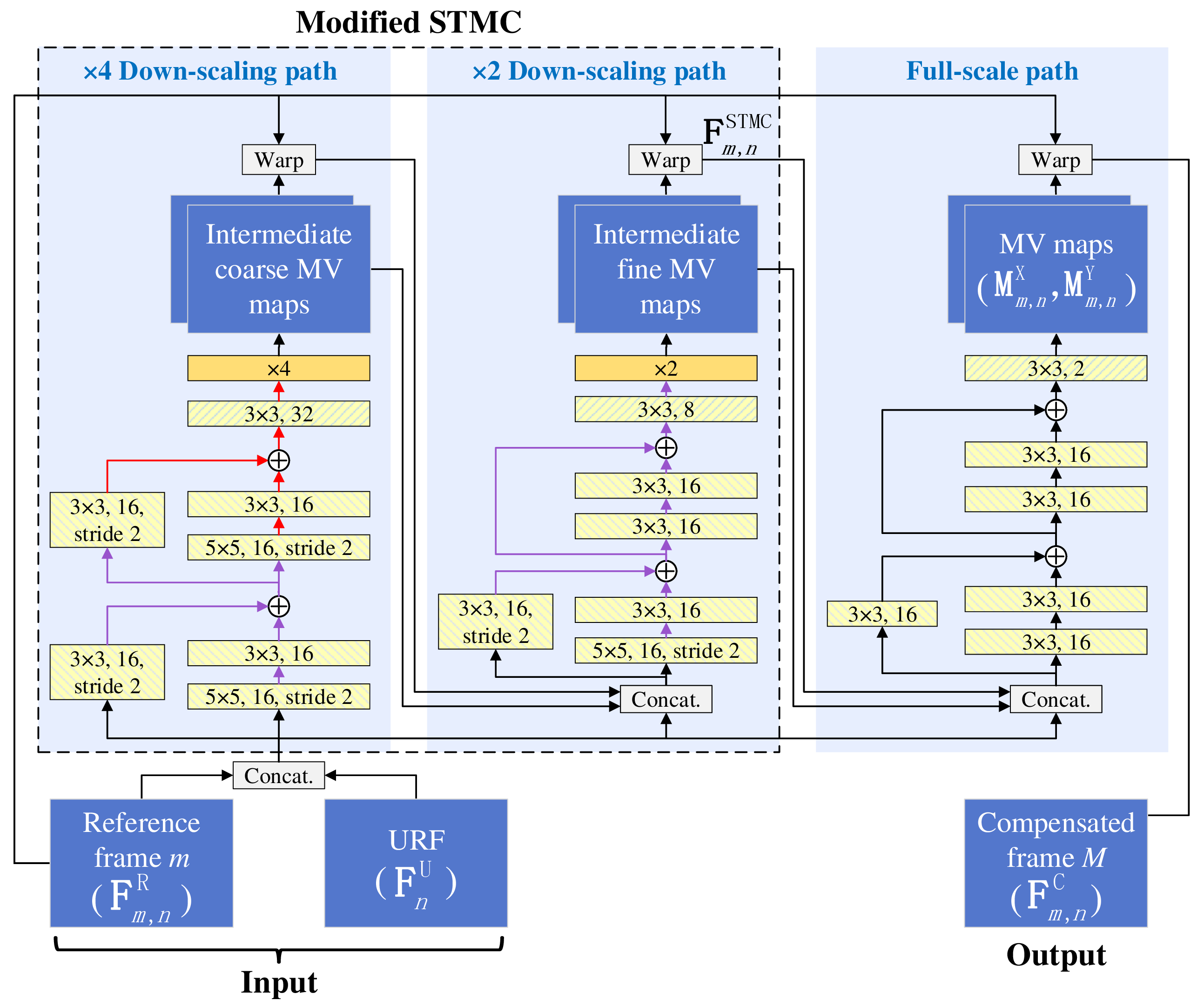}\\
		\vspace{-1.25em}
		\centerline{\hspace{7em}(a)}\medskip
	\end{minipage}
	\hfill
	\begin{minipage}{0.4\linewidth}
		\begin{minipage}{0.99\linewidth}
			\centering
			\includegraphics[width=36mm]{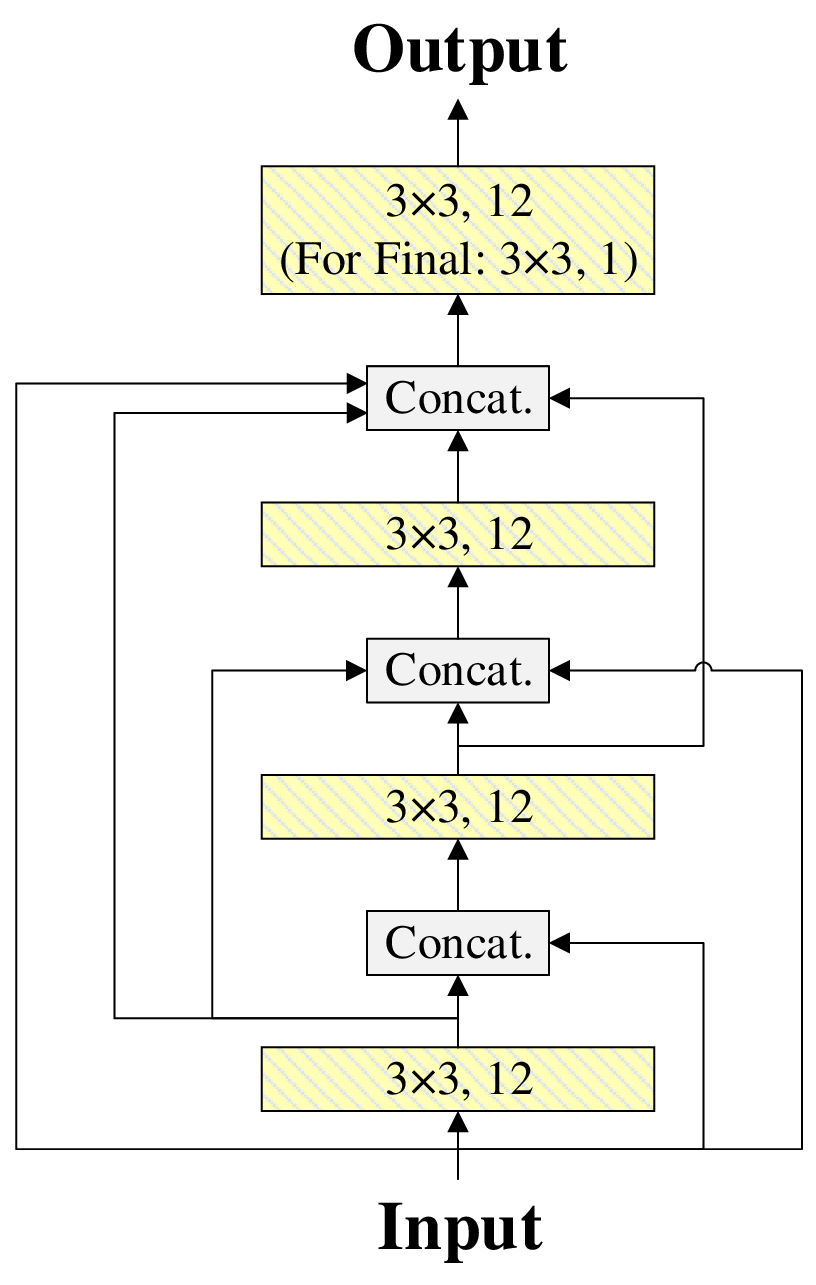}\\
			\centerline{(b)}\medskip
		\end{minipage}
		\vfill
		\begin{minipage}{0.99\linewidth}
			\centering
			\includegraphics[width=60mm]{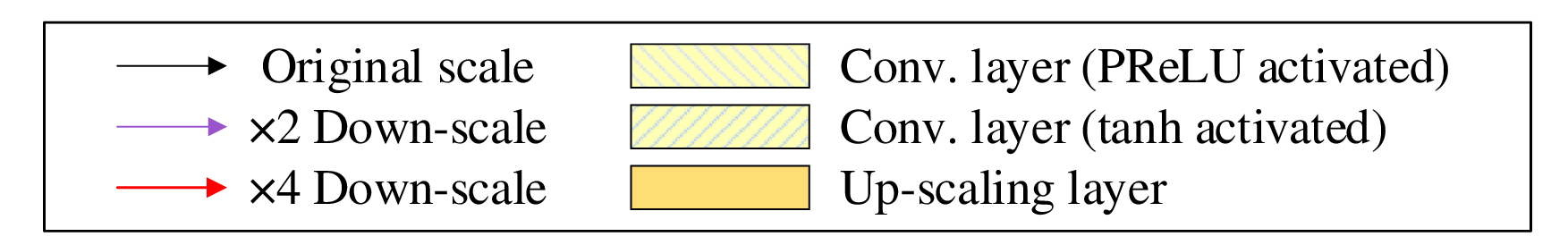}\\
		\end{minipage}
	\end{minipage}
	\vspace{-1.25em}
	\caption{Network details. (a) Motion compensation network. (b) Dense unit. 
		For convolutional layers, ``$p\times{}p$, $q$'' represents $q$ output channels with $p\times{}p$ kernels. Note that the convolutional stride is set to 1 by default, except that explicitly mentioned in certain layers.}
	\label{fig:net-detail}
	\vspace{-1.0em}
\end{figure}

\textbf{Motion compensation network.} We propose a motion compensation network based on the spatial transformer motion compensation (STMC) \cite{Caballero2017CVPR_STMC}, for content alignment between $\mathbf{F}^\mathrm{R}_{m,n}$ and $\mathbf{F}^\mathrm{U}_n$, illustrated in Figure \ref{fig:net-detail}-(a).
In \cite{Caballero2017CVPR_STMC}, the STMC takes both $\mathbf{F}^\mathrm{R}_{m,n}$ and $\mathbf{F}^\mathrm{U}_n$ as the input, to output a compensated frame denoted as $\mathbf{F}^\mathrm{STMC}_{m,n}$. 
The STMC consists of two paths ($\times$4 and $\times$2 down-scaling paths) to predict different precision of motion vector (MV) maps, and the MV maps from the $\times$2 down-scaling path are applied to $\mathbf{F}^\mathrm{R}_{m,n}$ for outputting $\mathbf{F}^\mathrm{STMC}_{m,n}$. 
The two down-sampling paths in \cite{Caballero2017CVPR_STMC} are capable for estimating various scales of motion. However, the accuracy of the STMC is limited due to down-sampling, and its architecture can also be improved. 
Therefore, we propose a motion compensation network with the following advancements.
(1) Besides the $\times$2 and $\times$4 down-scaling paths, a full-scale path is added to enhance the precision of MV estimation; (2) Inspired by the ResNet \cite{He2016CVPR_ResNet}, 6 shortcuts are added next to the convolutional layers for higher network capacity and ease to be trained; (3) All ReLU \cite{Glorot2011AISTATS_ReLU} activation for convolutional layers are replaced by PReLU \cite{He15ICCV_PReLU}.
With the above advancements, the full-scale path outputs two MV maps, $\mathbf{M}^\mathrm{X}_{m,n}$ and $\mathbf{M}^\mathrm{Y}_{m,n}$, denoting the horizontal and vertical motion of all pixels from $\mathbf{F}^\mathrm{R}_{m,n}$ to $\mathbf{F}^\mathrm{U}_n$. 
Finally, the compensated frame $\mathbf{F}^\mathrm{C}_{m,n}$ is derived by
\vspace{-0.5em}
\begin{equation}\label{eq:mc-interp}
	\mathbf{F}^\mathrm{C}_{m,n}(x,y)=\mathrm{Bil}\{\mathbf{F}^\mathrm{R}_{m,n}(x+\mathbf{M}^\mathrm{X}_{m,n}(x,y),y+\mathbf{M}^\mathrm{Y}_{m,n}(x,y))\},
	\vspace{-0.5em}
\end{equation}
where $x$ and $y$ are coordinates of a pixel, and $\mathrm{Bil}\{\cdot\}$ represents the bilinear interpolation considering that the motion may be of non-integer pixels.

\textbf{Block-adaptive convolutional layer.}
The input to this layer is a concatenation of three feature maps, including a compensated frame $\mathbf{F}^\mathrm{C}_{m,n}$, a URF $\mathbf{F}^\mathrm{U}_n$ and $\mathbf{F}^\mathrm{C}_{m,n}-\mathbf{F}^\mathrm{U}_n$. 
The CU and TU partition are represented by two feature maps, i.e., $\mathbf{C}_n$ and $\mathbf{T}_n$, respectively. 
$\mathbf{C}_n$ and $\mathbf{T}_n$ each has the same size as $\mathbf{F}^\mathrm{U}_n$, and the values in the two maps are assigned according to the partition structure. If pixel $(x,y)$ is on the boundary of a CU or TU, $\mathbf{C}_n(x,y)$ or $\mathbf{T}_n(x,y)$ is set to 1. Otherwise, the value is set to $-1$.
Afterwards, the target of this layer is to output a certain number of feature maps, providing three feature maps as the input and two feature maps as the guidance. 
For this problem, we present a guided convolution operation, assuming that $P^\mathrm{I}$, $P^\mathrm{G}$ and $P^\mathrm{O}$ feature maps are used as the input, guidance and output, respectively. 
The guided convolution consists of two main procedures, i.e., intermediate map generation and convolution with intermediation. First, the $P^\mathrm{G}$ guidance feature maps are processed with two typical convolutional layers to generate $P^\mathrm{O}$ intermediate feature maps, keeping the size of each feature map unchanged. 
Then, during the convolution, the $P^\mathrm{O}$ output feature maps are generated based on these $P^\mathrm{O}$ intermediate feature maps, correspondingly. Compared with typical convolution using space-irrelevant weights $w_{j,l}$ only, the guided convolution is conducted with space-relevant weights $w^\mathrm{G}_{j,l}$ generated from the intermediation, as formulated below
\vspace{-0.25em}
\begin{equation}\label{eq:conv-guided-1}
w^\mathrm{G}_{j,l}(\Delta{x},\Delta{y})=w_{j,l}(\Delta{x},\Delta{y}) \cdot  \mathbf{F}^\mathrm{M}_{l}(x+\Delta{x},y+\Delta{y}),\\
\vspace{-0.25em}
\end{equation}
\begin{equation}\label{eq:conv-guided-2}
\mathbf{F}^\mathrm{O}_{l}(x,y)=\sum_{j=1}^{P^\mathrm{I}}\sum_{\Delta{x}=-1}^1{}\sum_{\Delta{y}=-1}^1{}w^\mathrm{G}_{j,l}(\Delta{x},\Delta{y}) \cdot \mathbf{F}^\mathrm{I}_{j}(x+\Delta{x},y+\Delta{y}). \\  
\vspace{-0.25em}
\end{equation}
In (\ref{eq:conv-guided-1}) and (\ref{eq:conv-guided-2}), $\mathbf{F}^\mathrm{I}_{j}$, $\mathbf{F}^\mathrm{M}_{l}$ and $\mathbf{F}^\mathrm{O}_{l}$ represent the $j$-th input, the $l$-th intermediate and the $l$-th output feature maps, respectively. $\Delta{x},\Delta{y}$ denote the relative coordinates within a $3\times3$ kernel.
For each block-adaptive convolutional layer in MIF-Net, there exist $P^\mathrm{I}=3$ and $P^\mathrm{G}=2$, and we set the number of output maps to be $P^\mathrm{O}=16$.

\textbf{Dense units for quality enhancement.}
The DenseNet \cite{Huang2017CVPR_DenstNet} introduces various length of inter-layer connections, with alleviation of vanishing gradients and encouragement of feature reuse.
Considering the advantages, $(2M+2)$ dense units are adopted in MIF-Net, i.e., 2 dense units in each branch and 2 dense units at the end of MIF-Net synthesizing features from $M$ branches. Figure \ref{fig:net-detail}-(b) illustrates the structure of each dense unit, and it can be observed that a dense unit with 4 convolutional layers includes 10 inter-layer connections, much more than a 4-layer plain CNN with only 4 inter-layer connections. Here, each layer outputs 12 channels, except the last layer in the final dense unit outputting only 1 channel as the difference frame $\mathbf{F}^\mathrm{\Delta}_n$.

\textbf{MIF-Net Training.}
With both motion compensation and quality enhancement, it may be difficult to train the whole MIF-Net directly. Thus, we propose to train it with intermediate supervision \cite{Wei2016CVPR}, introducing two loss functions at different stages. First, the difference between $\mathbf{F}^\mathrm{U}_n$ and each frame in $\{\mathbf{F}^\mathrm{C}_{m,n}\}^M_{m=1}$ can measure the performance of motion compensation, and thus it is defined as the intermediate loss
\begin{equation}\label{eq:int-loss}
	L_{\mathrm{INT}}=\frac{1}{M}\sum_{m=1}^{M}\lVert{\mathbf{F}^\mathrm{C}_{m,n}-\mathbf{F}^\mathrm{U}_n}\rVert_2^2,
\end{equation}
where $\lVert{\cdot}\rVert_2$ represents the $\ell_2$-norm difference. 
Next, the difference between $\mathbf{F}^\mathrm{E}_n$ and $\mathbf{F}_n$ indicates the performance of the whole MIF-Net, and thus the global loss is
\begin{equation}\label{eq:glo-loss}
	L_{\mathrm{GLO}}=\lVert{\mathbf{F}^\mathrm{E}_n-\mathbf{F}_n}\rVert_2^2.
\end{equation}
The loss for training MIF-Net is the weighted summation of them:
\begin{equation}\label{eq:loss}
	L=\alpha\cdot{}L_{\mathrm{INT}}+\beta\cdot{}L_{\mathrm{GLO}},
\end{equation}
where $\alpha$ and $\beta$ are adjustable positive weights. 
On account that quality enhancement relies on the well-trained motion compensation network, $L_{\mathrm{INT}}$ should be emphatically optimized with $\alpha\gg\beta$ at early stage of training. After $L_{\mathrm{INT}}$ converges, we set $\beta\gg\alpha$ instead, to emphasize more on optimization of the global loss $L_{\mathrm{GLO}}$. 

\textbf{Difference between IF-Net and MIF-Net.} The difference between two networks lies in the absence of $M$ reference frames in IF-Net. Therefore, only quality enhancement without motion compensation is adopted in IF-Net, illustrated by red arrows in Figure \ref{fig:net}. Compared with MIF-Net, only one branch without any compensated frame exists in IF-Net, and the concatenation synthesizing $M$ branches is also omitted. Despite simpleness, a block-adaptive convolutional layer and four consecutive dense units still exist in IF-Net, ensuring sufficient network capacity. Considering no motion compensation in IF-Net, the loss of IF-Net is the same as $L_{\mathrm{GLO}}$ in MIF-Net. 

\vspace{-0.5em}
\section{Experimental Results}
\label{sec:result}
\vspace{-0.5em}

\vspace{-0.25em}
\subsection{Settings}
\label{sec:setting} 
\vspace{-0.25em}

\textbf{Experimental configurations.} 
In the experiments, all approaches for in-loop filtering were incorporated into the HEVC reference software HM 16.5.
For our MIF approach, we established a large-scale database for HEVC in-loop filtering (named HIF database) containing 111 raw video sequences, collected from the JCT-VC \cite{Ohm12TCSVT}, Xiph.org \cite{XIPH2017} and the conversational video set \cite{Xu2014JSTSP}. Our HIF database was divided into non-overlapping sets of training (83 sequences), validation (10 sequences) and test (18 sequences). The training set was used to train the networks, and the hyper-parameters in our approach were tunned on the validation set. The test set was used for performance evaluation, containing all 18 standard sequences from the JCT-VC set \cite{Ohm12TCSVT}. 
The RA configuration was applied for both network training and performance evaluation at four QPs, $\{22, 27, 32, 37\}$. 
During evaluation, the BD-BR and the Bj\o{}ntegaard delta PSNR (BD-PSNR) were measured to assess the rate-distortion (RD) performance.

\textbf{Network settings.}
For our approach, one MIF-Net model and one IF-Net model were trained for each evaluated QP, while all QPs shared the same trained RFS-Net model.
The tuned hyper-parameters for these networks are listed in Table \ref{tab:net-para}.
For training MIF-Net and IF-Net, all the frames were segmented into 64$\times$64 patches. Here, each training sample was composed of the co-located patches from a raw frame, a URF, a CU map, a TU map and $M$ reference frames (if have). Considering the efficiency of training, the IF-Net or MIF-Net model at QP $=37$ was trained from scratch, while the models at QPs $\{22, 27, 32\}$ were fine-tuned from the trained models at QPs $\{27, 32, 37\}$, respectively.

\begin{table}[t]
	\scriptsize
	\newcommand{\tabincell}[2]{\begin{tabular}{@{}#1@{}}#2\end{tabular}}
	\begin{center}
		\caption{Hyper-parameters for networks} \label{tab:net-para}
		\begin{tabular}{|c|c|c|}
			\hline Hyper-parameter & RFS-Net & MIF-Net or IF-Net \\
			\hline \tabincell{c}{Size of ref. frame pool: $N$} & 16 & - \\
			\hline \tabincell{c}{Threshold for CC value: $\tau$} & 0.3 & - \\
			\hline \tabincell{c}{Num. of selected ref. frames: $M$} & \multicolumn{2}{c|}{2} \\
			\hline \tabincell{c}{Optimization} & \multicolumn{2}{c|}{Adam algorithm \cite{Kingma2014CS_Adam}}\\
			\hline \tabincell{c}{Batch size} & $\le16^{^{*}}$ & 16 \\
			\hline \tabincell{c}{Initial learning rate} & $10^{-5}$ & $10^{-4}$ \\
			\hline \tabincell{c}{Num. of iterations} & $10^{5}$ & \tabincell{c}{$10^{6}$ (from scratch) or $2\times10^{5}$ (fine-tunning)} \\
			\hline \tabincell{c}{Changeable weights \\in MIF-Net: $\alpha$ and $\beta$} & - &\tabincell{c}{$0.99\:\&\:0.01$ (at beginning)\\ $0.01\:\&\:0.99$ (after $L_\mathrm{INT}$ converged)}\\
			
			\hline
		\end{tabular}
	\end{center}
    \vspace{-0.75em}
	\footnotesize{$^{*}$ The batch size equals to the number of valid reference frames for a URF.}
	\vspace{-0.5em}
\end{table}

\subsection{Performance Evaluation}
\label{sec:perform} 
\vspace{-0.25em}

\begin{table}
	\scriptsize
	\newcommand{\tabincell}[2]{\begin{tabular}{@{}#1@{}}#2\end{tabular}}
	\begin{center}
		\caption{\footnotesize{RD performance of in-loop filters on the JCT-VC test set}}
		\label{tab:rd-perform}
		
		\begin{tabular}{|c|c|c|c|c|c|c|c|c|c|}
			\hline \multirow{3}{*}{Class} & \multirow{3}{*}{\hms{}Sequence\hms{}} & \multicolumn{2}{c|}{\tabincell{c}{Standard\\DBF and SAO}} & \multicolumn{2}{c|}{\tabincell{c}{\hms{}Non-local adaptive\hms{}\\loop filter \cite{Zhang17TCSVT}}} & \multicolumn{2}{c|}{\tabincell{c}{RHCNN \cite{Zhang18TIP}}} & \multicolumn{2}{c|}{Proposed MIF} \\
			
			
			\cline{3-10} & & \hms{}BD-BR\hms{} & \hms{}BD-PSNR\hms{} & \hms{}BD-BR\hms{} & \hms{}BD-PSNR\hms{} & \hms{}BD-BR\hms{} & \hms{}BD-PSNR\hms{} & \hms{}BD-BR\hms{} & \hms{}BD-PSNR\hms{} \\
			& & (\%) & (dB) & (\%) & (dB) & (\%) & (dB) & (\%) & (dB) \\
			
			\hline \multirow{2}{*}{A} & \textit{\hms PeopleOnStreet \hms} & -8.29 & 0.37 & -12.03 &  0.54  & -12.48 & 0.57 & \textbf{-16.82} & \textbf{0.78} \\
			\cline {2-10} & \textit{\hms Traffic \hms}                    & -5.35 & 0.16 & -6.17 &  0.19 & -9.81 & 0.30 & \textbf{-12.15} & \textbf{0.38} \\
			
			\hline \multirow{5}{*}{B} & \textit{\hms BasketballDrive } & -6.65 & 0.15 & -8.84 & 0.20 & -11.05 & 0.25 & \textbf{-14.87} & \textbf{0.35} \\
			\cline {2-10} & \textit{\hms BQTerrace }                   & -7.15 & 0.11 & -11.40 & 0.17 & -14.36 & 0.23 & \textbf{-17.13} & \textbf{0.27} \\
			\cline {2-10} & \textit{\hms Cactus \hms}                  & -7.54 & 0.16 & -8.90  & 0.19 & -12.52 & 0.27 & \textbf{-15.83} & \textbf{0.35} \\
			\cline {2-10} & \textit{\hms Kimono \hms}                  & -7.54 & 0.22 & -9.26 &  0.27 & -10.48 & 0.31 & \textbf{-12.24} & \textbf{0.37} \\
			\cline {2-10} & \textit{\hms ParkScene \hms}               & -3.68 & 0.11 & -4.08 &  0.12 & -5.94 & 0.18 & \textbf{-7.99} & \textbf{0.25} \\
			
			\hline \multirow{4}{*}{C} & \textit{\hms BasketballDrill \hms} & -5.02 & 0.21 & -5.39 &  0.22 & -7.81 & 0.33 & \textbf{-10.32} & \textbf{0.43} \\
			\cline {2-10} & \textit{\hms BQMall \hms}                      & -3.93 & 0.15 & -4.45 &  0.17 & -7.65 & 0.30 & \textbf{-9.38} & \textbf{0.37} \\
			\cline {2-10} & \textit{\hms PartyScene \hms}                  & -1.05 & 0.04 & -1.22 &  0.05 & -2.41 & 0.10 & \textbf{-4.16} & \textbf{0.17} \\
			\cline {2-10} & \textit{\hms RaceHorses \hms}                  & -6.15 & 0.22 & -7.08 &  0.26 & -10.40 & 0.39 & \textbf{-12.74} & \textbf{0.48} \\
			
			\hline \multirow{4}{*}{D} & \textit{\hms BasketballPass \hms} & -3.85 & 0.18 & -4.32 &  0.20 & -7.68 & 0.37 & \textbf{-9.98} & \textbf{0.48} \\
			\cline {2-10} & \textit{\hms BlowingBubbles \hms}             & -0.83 & 0.03 & -0.83 &  0.03 & -3.05 & 0.12 & \textbf{-3.98} & \textbf{0.16} \\
			\cline {2-10} & \textit{\hms BQSquare \hms}                   & -0.05 & 0.00 & 0.01 &  0.00 & -3.24 & 0.12 & \textbf{-4.40} & \textbf{0.17} \\
			\cline {2-10} & \textit{\hms RaceHorses \hms}                 & -4.44 & 0.20 & -4.80 &  0.22 & -8.84 & 0.41 & \textbf{-10.99} & \textbf{0.51} \\
			
			\hline \multirow{3}{*}{E} & \textit{\hms FourPeople \hms} & -7.02 & 0.26 & -8.49 & 0.32 & -13.92 & 0.54 & \textbf{-16.48} & \textbf{0.64} \\
			\cline {2-10} & \textit{\hms Johnny \hms}                 & -5.60 & 0.14 & -8.03 & 0.21 & -11.62 & 0.30 & \textbf{-14.37} & \textbf{0.38} \\
			\cline {2-10} & \textit{\hms KristenAndSara \hms}         & -6.41 & 0.20 & -8.01 & 0.25 & -12.62 & 0.41 & \textbf{-15.34} & \textbf{0.50} \\
			
			\hline \multicolumn{2}{|c|}{Average}                 & -5.03 & 0.16 & -6.29 & 0.20 & -9.22 & 0.30 & \textbf{-11.62} & \textbf{0.39} \\
			\hline 
			
		\end{tabular}
	\end{center}
	\vspace{-2em}
\end{table}

\textbf{Objective RD performance.} We analyze the objective performance of our MIF approach in terms of the BD-BR and BD-PSNR, compared with the standard in-loop filters (DBF and SAO), a model-based approach (the non-local adaptive loop filter \cite{Zhang17TCSVT}) and a deep-learning-based approach (the RHCNN \cite{Zhang18TIP}). For a fair comparison, the RHCNN models in \cite{Zhang18TIP} were re-trained on our HIF database. Table \ref{tab:rd-perform} tabulates the RD performance of all four approaches, and the original HM without in-loop filters is used as anchor. As indicated in Table \ref{tab:rd-perform}, the BD-BR of our MIF approach is $-11.62\%$ averaged over the 18 standard test sequences, outperforming $-5.03\%$ of the HM baseline, $-6.29\%$ of \cite{Zhang17TCSVT} and $-9.22\%$ of \cite{Zhang18TIP}. 
In terms of BD-PSNR, our approach achieves $0.39$dB for the standard test set, also significantly better than $0.16$dB of the HM baseline, $0.20$dB of \cite{Zhang17TCSVT} and $0.30$dB of \cite{Zhang18TIP}, respectively. Therefore, our MIF approach achieves the best RD performance among all four approaches.  
The advancement of our approach mainly attributes to the accurate mapping from a URF to its corresponding raw frame, benefiting from the deep MIF-Net and IF-Net learned on our large-scale HIF database.

\begin{figure*}
	\centering
	\includegraphics[width=1.0\linewidth]{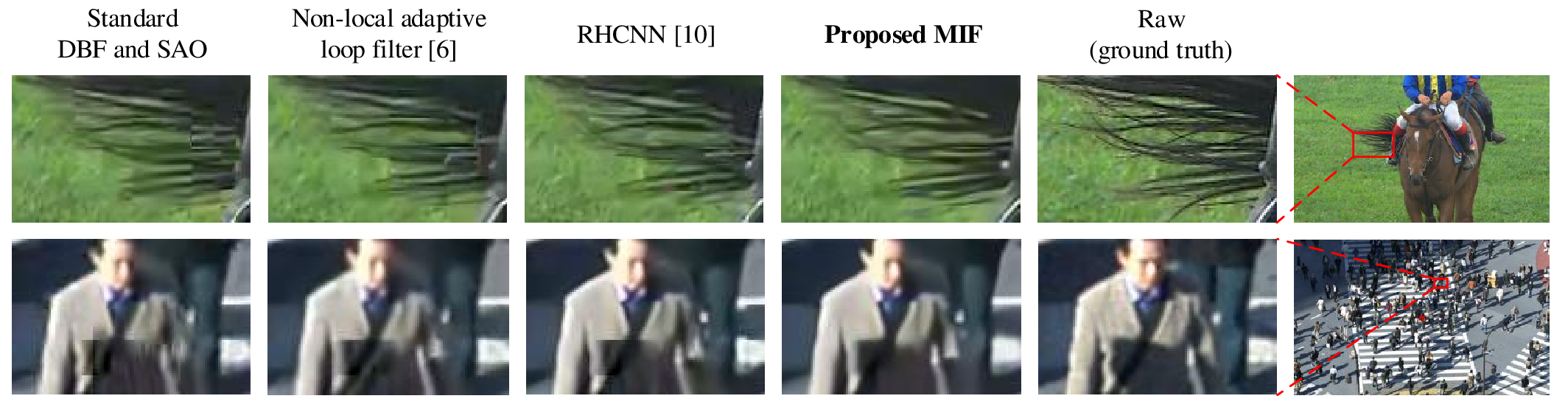}
	\vspace{-2.25em}
	\caption{Comparison of subjective visual quality on sequences \textit{RaceHorses} (Class C) and \textit{PeopleOnStreet} (Class A) at QP $=37$.}
	\label{fig:subj-quality}
	\vspace{-1em}
\end{figure*}

\textbf{Subjective visual quality.}
Figure \ref{fig:subj-quality} illustrates the subjective visual quality among all four approaches. It can be observed that the frames enhanced by our approach remain less distortion than those by other approaches, e.g., the clearer edge of the horse tail and the reduced blocking artifacts on the pedestrians. The highest visual quality mainly benefits from the utilization of multiple adjacent frames in the proposed MIF approach.

\vspace{-0.5em}
\section{Conclusion}
\label{sec:conclusion}
\vspace{-0.5em}

In this paper, we have proposed a DenseNet based in-loop filter for HEVC. Different from existing in-loop filter approaches based on a single frame, our MIF approach enhances the quality of each encoded frame leveraging multiple adjacent frames. To this end, we first propose an RFS to find higher-quality frames. Then, we develop an MIF-Net model for multi-frame in-loop filter in HEVC, which is based on the DenseNet and benefits from the improved generalization capacity and computational efficiency. Finally, experimental results demonstrate that our approach achieves $-11.62\%$ of BD-BR saving and $0.39$dB of BD-PSNR increment on average, outperforming the HM baseline and other state-of-the-art approaches.

\vspace{0.5em}
\Section{Acknowledgment}
\label{sec:ack}
\vspace{-0.5em}

This work was supported by NSFC under Grants 61876013 and 61573037, and by the Fok Ying
Tung Education Foundation under Grant 151061.

\Section{References}
\bibliographystyle{IEEEbib}
\bibliography{refs}

\end{document}